\documentclass{article}

\usepackage{arxiv}

\usepackage[utf8]{inputenc} 
\usepackage[T1]{fontenc}    
\usepackage{hyperref}       
\usepackage{url}            
\usepackage{booktabs}       
\usepackage{amsfonts}       
\usepackage{nicefrac}       
\usepackage{microtype}      

\usepackage{graphicx}
\usepackage{algorithm}
\usepackage{algpseudocode}
\usepackage{xcolor}
\usepackage{amssymb}
\usepackage{subcaption}
\usepackage{booktabs}
\usepackage{enumerate}
\usepackage{amsmath}
\usepackage{appendix}

\usepackage{booktabs}
\usepackage{amsfonts}
\usepackage{amsmath}
\usepackage{times}  
\usepackage{helvet} 
\usepackage{courier}  
\usepackage{graphicx}  
\usepackage{grffile}
\title{Bayesian Neural Decoding Using A Diversity-Encouraging Latent Representation Learning Method}

\author{
  Tian Chen \\
  Department of Statistics\\
  UC Irvine, CA, USA\\
   \And
 Lingge Li \\
Department of Statistics\\
  UC Irvine, CA, USA\\
  \AND
  Gabriel Elias \\
  Department of Neurobiology and Behavior \\
  UC Irvine, CA, USA \\
  \AND
  Norbert Fortin \\
  Department of Neurobiology and Behavior \\
  UC Irvine, CA, USA \\
  \And
  Babak Shahbaba\thanks{babaks@uci.edu} \\
  Department of Statistics\\
  UC Irvine, CA, USA\\
}

\begin{document}
\maketitle

\begin{abstract}
It is well established that temporal organization is critical to memory, and that the ability to temporally organize information is fundamental to many perceptual, cognitive, and motor processes. While our understanding of how the brain processes the spatial context of memories has advanced considerably, our understanding of their temporal organization lags far behind. In this paper, we propose a new approach for elucidating the neural basis of complex behaviors and temporal organization of memories. More specifically, we focus on neural decoding --- the prediction of behavioral or experimental conditions based on observed neural data. In general, this is a challenging classification problem, which is of immense interest in neuroscience. Our goal is to develop a new framework that not only improves the overall accuracy of decoding, but also provides a clear latent representation of the decoding process. To accomplish this, our approach uses a Variational Auto-encoder (VAE) model with a diversity-encouraging prior based on determinantal point processes (DPP) to improve latent representation learning by avoiding redundancy in the latent space. We apply our method to data collected from a novel rat experiment that involves presenting repeated sequences of odors at a single port and testing the rats' ability to identify each odor. We show that our method leads to substantially higher accuracy rate for neural decoding and allows to discover novel biological phenomena by providing a clear latent representation of the decoding process.
\end{abstract}

\section{Introduction}
Our main focus in this paper is to elucidate the underlying neuronal mechanisms for the memory of sequence of events. There is considerable evidence that hippocampal neurons can encode sequences of locations \cite{skaggs1996replay,mehta2000experience,dragoi06,foster2006reverse,gupta2012segmentation}. However, there is no direct evidence that this coding property extends to nonspatial memories (the type of memory typically investigated in human studies). Recent electrophysiological studies have specifically targeted this important issue \cite{eichenbaum14,allen14,allen15,allen2016nonspatial}. These studies commonly involve recording spike train data, which are sequences of spikes (action potentials) produced by neurons over time \cite{mitzdorf1985current}. 

\begin{figure*}[ht]
\centering
\includegraphics[width=1.0\linewidth]{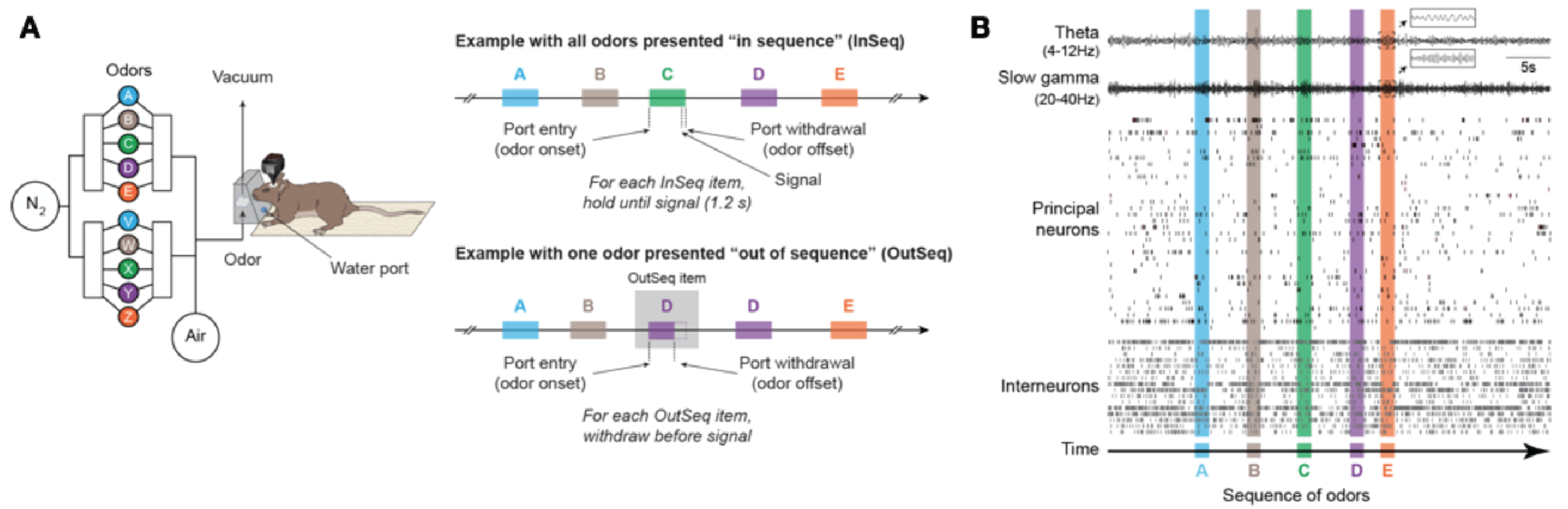}
\caption{\textbf{Nonspatial sequence memory task.}
Neural activity was recorded as rats performed a cross-species sequence memory, which shows strong
behavioral parallels in rats and humans. Briefly, this hippocampus-dependent task involves repeated presentations of sequences of nonspatial items (odors) and requires subjects to determine whether each item is presented ``in sequence'' (InSeq) or ``out of seq
uence'' (OutSeq). \textbf{A}) Using
an automated odor delivery system, rats were presented with series of five odors delivered in the same odor port. In each session, the same sequence was presented multiple times, with approximately half the presentations including all items InSeq (ABCDE) and the other half including one item OutSeq (e.g., ABDDE). Each odor presentation was initiated by a nosepoke and rats were required to correctly identify the odor as either InSeq (by holding their nosepoke response until the signal at 1.2s) or OutSeq (by withdrawing their nosepoke before the signal; $<1.2$s) to receive a reward. \textbf{B}) Example activity from simultaneously recorded CA1 neurons (bottom) and local field potentials oscillations (LFP; theta and beta/slow-gamma bands; top) during one sequence presentation. }
\label{fig:Task}
\end{figure*}

To study temporal organization of memory, we have designed an experiment to record neural activity in the CA1 region of the hippocampus as rats performed a nonspatial sequential memory task (see Figure \ref{fig:Task} for more details). The task involves the presentation of repeated sequences of odors at a single port and tests the rats' ability to identify each odor as ``in sequence'' (InSeq; by holding their nosepoke response until the signal at 1.2s) or ``out of sequence'' (OutSeq; by withdrawing their nose before the signal). Spiking activities were recorded using 24 tetrodes (bundles of 4 electrodes) in rats tested on well-trained or novel sequences of 5 odors denoted as A, B, C, D, and E. For each odor presentation (trial), the data typically features spike counts from $\sim$40-70 neurons and trial identifiers (e.g., Odor presented, InSeq/OutSeq, response correct/incorrect). Here, our goal is to identify the presented odors (each treated as a class) given the spike train data. 


Properly analyzing the massive amount of data collected through this experiment is an extremely difficult task. Many existing statistical methods and machine learning techniques lack the flexibility, rigor, and scalability required for analyzing such complex and high dimensional data. What makes this task more daunting is the fact that the underlying patterns tend to be highly nonlinear. Moreover, the patterns are typically dominated by a specific behaviour or experimental condition mainly due to the class imbalance problem in the data. In our case, the patterns are dominated by odor A since it is the first odor in the sequence, is always in the right order (i.e., always occurs first), and is included in all trials. In contrast, odors D and E might not appear in some trials if the trials end early due to rats' mistakes in recognizing earlier odors.  

In general, class imbalance is a commonly encountered problem in many real-world problems, where some classes might be infrequent or rare so they are not represented properly in finite samples. As a result, the analysis might lead to biased estimates and inaccurate predictions. In supervised learning, the classification model might be dominated by the most frequent classes ignoring infrequent, but usually important classes. Imbalance can also affect unsupervised learning models. For example, for latent variable models, the low dimensional representation of the data might be mainly dominated by the information collected on more frequently observed classes. 



In this paper, we propose a framework that can be used to explore, visualize, describe, and predict neural patterns. More specifically, we propose a method for alleviating the class imbalance problem in latent variable models to improve the overall quality of latent representation learning and also to increase the accuracy of the decoding process. To this end, a diversity-encouraging prior, namely determinantal point process (DPP), is used for latent variables. The prior plays the role of regularizing the latent variables to be less redundant. In particular, we use this prior for the latent variables in Variational Auto-encoder (VAE), which is one of the most widely used deep latent model. Compared to the standard VAE models, we expect our method to provide a more clear latent representation of the data by avoiding redundancy in the latent space. 

Our proposed method has the following advantages: 1) it can capture nonlinear patterns and map the observed data into a latent spaces where the linearity assumption (in terms of the relationship between neural data and behaviour) would be more reasonable; 2) by using a DPP prior, our method has a better chance to separate different classes in order to provide a more clear representation of the data; 3) additionally, our approach can ensure that the patterns associated with minor classes (here, odors D and E) are well represented in the latent space and are decoded more accurately; 4) finally, our proposed method can provide unprecedented insight into the hippocampal mechanisms underlying the memory for sequences of events (a defining feature of episodic memory) and can also determine whether spatial coding properties (e.g., sequence replay) thought to be fundamental to hippocampal function extend to the nonspatial domain. 

The paper is organized as follows. First, we describe the details of the experiment that inspired this research. We then provide a brief overview of Variational Auto-encoder and Determinantal Point Processes. We also review some existing methods for dealing with class imbalance. Next, we describe our proposed method in details. Finally, we provide our analyses and results for two separate problems. First, we show the advantages of our method based on an imbalanced MNIST classification task. We then use our method for analyzing the data collected from the rat experiment, which is a multi-class and high-dimensional neurophysiological experiment. 



\section{Experimental design and data collection}\label{experiment}

Subjects were male Long-Evans rats, weighing $\sim$350g at the beginning of the experiment. They were individually housed and maintained on a 12h light/dark cycle. Rats had ad libitum access to food and their hydration levels were monitored daily. All procedures were conducted in accordance with the Institutional Animal Care and Use Committee. 

The apparatus consisted of a linear track with walls angled outward. An odor port, located on one end of the track, was equipped with photobeam sensors to precisely detect nose entries and was connected to an automated odor delivery system capable of repeated deliveries of multiple distinct odors. Two water ports were used for reward delivery: one located under the odor port, the other at the opposite end of the track. Timing boards (Plexon) and digital input/output boards (National Instruments) were used to measure response times and control the hardware. All aspects of the task were automated using custom Matlab scripts (MathWorks). A 96-channel Multichannel Acquisition Processor (MAP; Plexon) was used to interface with the hardware in real time and record the behavioral and electrophysiological data. Odor stimuli consisted of synthetic food extracts contained in glass jars (A, lemon; B, rum; C, anise; D, vanilla; E, banana) that were volatilized with desiccated, charcoal-filtered air (flow rate, 2 L/min). To prevent cross-contamination, separate Teflon tubing lines were used for each odor. These lines converged in a single channel at the bottom of the odor port. In addition, an air vacuum located at the top of the odor port provided constant negative pressure to quickly evacuate odor traces. Readings from a volatile organic compound detector confirmed that odors were cleared from the port 500-750 ms after odor delivery (inter-odor intervals were limited by software to $\ge$800 ms).

Na\"{i}ve rats were initially trained to nosepoke and reliably hold their nose for 1.2 sec in the odor port for a water reward. Odor sequences of increasing length were then introduced in successive stages (A, AB, ABC, ABCD, and ABCDE) upon reaching behavioral criterion of 80\% correct over three sessions per training stage. In each stage, rats were trained to correctly identify each presented item as either InSeq (by holding their nosepoke response for $>$1.2 sec to receive reward) or OutSeq (by withdrawing their nose before 1.2 sec to receive reward). There were two types of OutSeq items in the dataset, Repeats, in which an earlier item was presented a second time in the sequence (e.g., ABADE), and Skips, in which an item was presented too early in the sequence (e.g., ABDDE, which skipped over item C). Although our previous work has revealed important differences in performance and neural activity on Repeats and Skips, this distinction was beyond the scope of the present analyses and not further discussed here. Note that OutSeq items could be presented in any sequence position except the first (i.e., sequences always began with odor A, though odor A could also be presented later in the sequence as a Repeat). After reaching criterion performance on the five-item sequence ($>$80\% correct on both InSeq and OutSeq items), rats underwent microdrive implantation.

Each chronically implanted microdrive contained 20 independently drivable tetrodes. Voltage signals recorded from the tetrode tips were referenced to a ground screw positioned over the cerebellum, and filtered for single-unit activity (154 Hz to 8.8 kHz). The neural signals were then amplified (10,000-32,000), digitized (40 kHz) and recorded to disk with the data acquisition system (MAP, Plexon). Action potentials from individual neurons were manually isolated off-line using a combination of standard waveform features across the four channels of each tetrode (Offline Sorter, Plexon). Proper isolation was verified using interspike interval distributions for each isolated unit (assuming a minimum refractory period of 1 ms) and cross-correlograms for each pair of simultaneously recorded units on the same tetrode. To confirm recording sites, current was passed through the electrodes before perfusion (0.9\% PBS followed by 4\% paraformaldehyde) to produce small marking lesions, which were subsequently localized on Nissl-stained tissue slices.

\section{Background}
\label{relatedWork}

\subsection{Class imbalance}
As summarized in \cite{he2009learning}, there are two major branches of methods for dealing with imbalanced learning problem: sampling (data-level) methods and cost-sensitive (model-level) methods. The goal of imbalanced learning is usually to improve the performance of the underrepresented (minnor) classes, sometimes at the cost of slightly worse performance for the dominating classes.

Random oversampling and undersampling are the most popular sampling methods since they are easy and straightforward to implement. However, undersampling has the problem of losing important information regarding the major class, while oversampling might result in overfitting \cite{holte1989concept,mease2007boosted}.

Cost-sensitive methods usually associate a higher cost with misclassifying minor classes compared to misclassifying major classes \cite{elkan2001foundations}. For these model-specific methods, several algorithms based on cost-sensitive boosting, decision trees, and neural networks have been developed \cite{he2009learning}. In practice, however, finding a reasonable cost function is not trivial. 

In contrast to these alternatives, our proposed method in this paper takes a probabilistic perspective. Instead of `up-weighting' the cost of misclassifying minor classes, our method intrinsically `up-weight' them in the prior assumed for the latent representation by using DPP (this is illustrated in the Results section). Additionally, while most existing methods designed for class imbalance focus on binary classification problems, our approach can be easily applied to general classification problems with multiple classes. 

\subsection{Variational auto-encoder}
Variational auto-encoder (VAE) \cite{kingma2013auto} is an unsupervised model for learning low-dimensional latent representation of a given dataset. It can also be used for learning deep generative models to generate realistic data such as images and texts. Compared to a standard auto-encoder, variational auto-encoder imposes a prior on the latent variable instead of treating it as a deterministic term. The prior can also be regarded as a regularizor on the latent variablez.

More formally, we are interested in a dataset, $X$, and its latent representation $Z$. In the framework of graphical models, latent data $z$ is generated from a pior $p_{\theta}(z)$, and data $x$ is generated from the model $p_{\theta}(x \vert z)$. VAE aims at efficient marginal inference for $x$ as well as finding an approximation for the posterior distribution of $z$ using a simplified model $q_{\phi}(z \vert x)$ \cite{kingma2013auto}. Unlike mean-field variational Bayes, VAE simultaneously maximizes marginal likelihood of $x$ (evidence) and minimizes KL-divergence (KLD) between $q_{\phi}(z \vert x)$ and $p_{\theta}(z \vert x)$. This is achieved by maximizing the evidence lower bound (ELBO) with respect to $\theta$ and $\phi$: \begin{align*}
\mathcal{L}(\theta, \phi; x) & = \log p_{\theta}(X) - KL(q_{\phi}(z \vert x) \Vert p_{\theta}(z \vert x) ) \\
&  = E_{q_{\phi}(z \vert x)}(\log p_{\theta}(x \vert z)) - KL(q_{\phi}(z \vert x) \Vert p_{\theta}(z) )
\end{align*} 
This is equivalent to minimizing $- E_{q_{\phi}(z \vert x)}(\log p_{\theta}(x \vert z)) $, called the \emph{reconstruction loss}, while minimizing KLD between $q_{\phi}(z \vert x)$ and $p_{\theta}(z)$ at the same time.

The basic steps of a VAE involves passing observed data, $x$, into an encoder model to learn $q_{\phi}(z \vert x)$. Then, latent variable $z$ is sampled from $q_{\phi}(z \vert x)$ and fed into a decoder model to learn $p_{\theta}(x \vert z)$, which can be used to reconstruct new $x$. VAE involves learning both the inference model $q_{\phi}(z \vert x)$ and the generative model $p_{\theta}(x \vert z)$. The inference model $q_{\phi}(z \vert x)$, which called the encoder, is a posterior distribution of the latent variable given the data. Meanwhile, the  generative model $p_{\theta}(x \vert z)$, which is called the decoder, maps latent variables to a generative distribution for $x$. Both decoder and encoder models can be approximated by neural networks, which is capable of learning most of the complex and nonlinear patterns.


\subsection{Determinantal Point Process (DPP) and k-DPP}
DPP is a point process for modeling repulsion interactions between samples \cite{kulesza2012determinantal}. It defines a distribution over subsets of a fixed ground set, and assigns higher probability to more diverse subsets. In a discrete setting, let us denote the ground set as $\mathcal{Y}$. Then, $\mathcal{P}$ is defined to be a determinantal point process if for every $A \subseteq \mathcal{Y}$ we have 
\begin{align*}
\mathcal{P}(A \subseteq Y) \propto det(L_A)
\end{align*}
where Y is a subset randomly drawn from $\mathcal{Y}$ according to $\mathcal{P}$. $L$ is a kernel matrix: $\mathcal{Y} \times \mathcal{Y}   \rightarrow \mathbb{R} $, and $L_A$ is its submatrix corresponding to all entries in $A$. For example, if $A = \{ i, j\}$, where $i, j \in \mathcal{Y}$, then:
\begin{align*}
\mathcal{P}(A \subseteq Y) \propto det(L_A) & =  \begin{vmatrix}
L_{ii}& L_{ij}\\ 
L_{ji} & L_{jj} 
\end{vmatrix}
\end{align*}
Note that because the likelihood is proportional to the determinant of $L_A$, as a result, it is also proportional to the square of the volume spanned by the element vectors. Therefore, the likelihood becomes smaller for subsets with similar elements.

In a continuous setting, denote the ground set as $\Omega \subseteq \mathbb{R}^D$. Similar to the discrete set, we have a positive definite kernel function $L: \Omega \times \Omega \rightarrow \mathbb{R}$. For any $A \subseteq  \Omega$, we have $P_L(A) \propto \det(L_A)$.

In some cases, it is necessary to fix the subset size for every draw. A determinantal point process over subsets with cardinality $k$ is denoted as $k$-DPP. For the discrete setting, the likelihood is: 
\begin{align*}
P_L(A) = \dfrac{det(L_A)}{\sum \limits_{\vert B \vert = k} det(L_B)} = \dfrac{det(L_A)}{e_k(\lambda_1, \cdots, \lambda_N)}
\end{align*}
where $\lambda_1, \cdots, \lambda_N$ are eigenvalues of $L$, and $e_k(\lambda_1, \cdots, \lambda_N)$ is the $k$th elementary symmetric polynomial \cite{affandi2014learning}. Similarly, for the continuous setting we have
\begin{align*}
P_L(A) =  \dfrac{det(L_A)}{e_k(\lambda_{1:\infty})}
\end{align*}
However, the term $e_k(\lambda_{1:\infty})$ is generally infeasible to evaluate.

\section{Methods}
\label{methods}

\subsection{Diverse latent variables using a k-DPP prior}
In the original VAE model, a standard normal prior is used for the latent variable $z$. Instead, we propose to use a continuous k-DPP prior for the continuous latent space, with the cardinality set to be the sample size $N$. This way, we have the following prior:
$$p_{\theta}(z) = \mathcal{P}_L^N( z)  = \dfrac{det(L_Z)}{e_N(\lambda_{1:\infty})} $$
where $L_Z$ is a $N \times N $ kernel matrix. 

Notice that the loss function of VAE is composed of the reconstruction loss and the KLD loss. For DPP-VAE, the reconstruction loss remains the same, while the KLD loss is modified to be
\begin{flalign*}
	KL(q_{\phi}(z \vert x) \Vert p_{\theta}(z) )  & =   \sum \limits_{n=1}^N(-\log \vert \Sigma \vert - P)\\
	& -  E_{q_{\phi}(z \vert x)}(\log p_{\theta}(z))\\
	& \cong  \sum \limits_{n=1}^N(-\log \vert \Sigma \vert  P)\\
	& -  \ln \det (L_Z)  + \ln(e_N(\lambda_{1:\infty})
\end{flalign*}
That is, to avoid large penalty from the KLD loss, the approximating distribution $q_{\phi}(z \vert x)$ needs to generate more diverse (hence, more balanced) latent variables $z$ across all classes. This is due to the fact that samples from the same class tend to be more similar compared to samples from different classes. As a result, when most samples are from the same class (more likely from the dominating class), the term $\det(L_Z)$ will become smaller. 

\subsection{Inference}
\begin{figure}[t]
\centering
\includegraphics[width=0.7\linewidth]{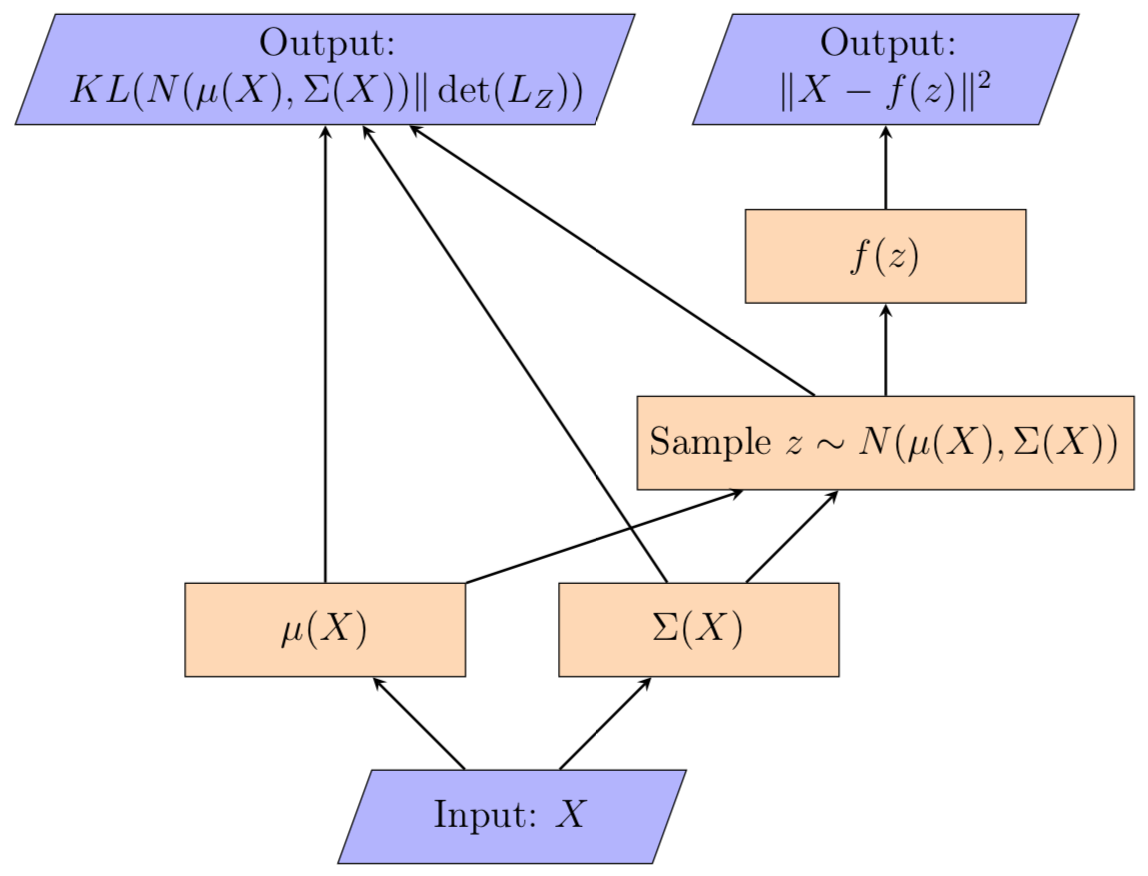}
\caption{DPP-VAE Architecture assuming $X$ is continuous.}
\label{fig:dppvae_architecture}
\end{figure}

The other components of our model remain similar to the conventional VAE. The approximate posterior distribution for latent variable is modeled as a normal distribution: $q_{\phi}(z \vert x) = N(z \vert \mu (X), \Sigma (X))$. The generative model can be written as $p_{\theta}(x \vert z) = p_{\theta}(x \vert f(z))$. For example, if $x$ is continuous, we can use $p_{\theta}(x \vert z) = N(x \vert f(z), I)$. If $x$ is binary, we have $p_{\theta}(x \vert z) = Bern(x \vert f(z))$. Note that $\mu (X)$, $\Sigma (X)$ and $f(z)$ are modeled by neural networks as nonlinear functions of ($X, \phi$) and ($Z, \theta$), where $\phi$ and $\theta$ are the weight vectors. The neural network structure of our method is shown in Figure \ref{fig:dppvae_architecture}.

Similar to obtaining the reconstruction loss, we use Monte Carlo estimates of the expectation of $\log p_{\theta}(z)$ with respect to
$q_{\phi}(z \vert x)$. More specifically, we choose a positive definite kernel function $ {L(X)}_{nm}  = q(\mathbf{x}_n)k(\mathbf{x}_n, \mathbf{x}_m) q(\mathbf{x}_m)$ as suggested by \cite{affandi2014learning,fasshauer2012stable}, where 
\begin{align*}
q(x) & = \sqrt{\alpha} \prod \limits_{d=1}^D \dfrac{1}{\sqrt{\pi \rho_d}} \exp(-\dfrac{x_d^2}{2\rho_d}) \\
k(x,y) &=  \prod \limits_{d=1}^D  \exp(-\dfrac{(x_d-y_d)^2}{2\sigma_d}) 
\end{align*}
The eigenvalues can be obtained by
$$\lambda_n = \alpha  \prod \limits_{d=1}^D (\dfrac{\beta_d^2+1}{2}+\dfrac{1}{2\gamma_d})^{-\frac{1}{2}}(\gamma_d(\beta_d^2 +1)+1)^{1-n_d} $$

Note that the term $ e_k(\lambda_{1:\infty})$ in the KLD loss has no explicit form, but as shown in \cite{affandi2014learning}, it has the following lower and upper bounds:
\begin{align*}
e_k(\lambda_{1:M})  & <=  e_k(\lambda_{1:\infty}) \\
 & <= \sum \limits_{j=0}^k \dfrac{(tr(L) - \sum^M_{n=1} \lambda_n)^j}{j!} e_{k-j}(\lambda_{1:M})\\
\end{align*}
To keep the KLD loss positive, we can use the upper bound of the normalization term for approximation. Moreover, $e_k(\lambda_{1:M})$ can be efficiently computed by the algorithm developed in \cite{kulesza2012determinantal}.

\section{Results}\label{results}
In this section, we apply our proposed method, henceforth called DPP-VAE, to two different problems: 1) MNIST classification, and 2) Bayesian neural decoding of the odor experiment discussed in the introduction section. 

\subsection{Two-class MNIST data}
In this example, VAE and DPP-VAE are trained to draw MNIST digits. We show the effectiveness of DPP-VAE in terms of balancing the class ratios of the generated data and classification accuracy.

\begin{figure}[!t]
\centering
\includegraphics[width=0.5\textwidth]{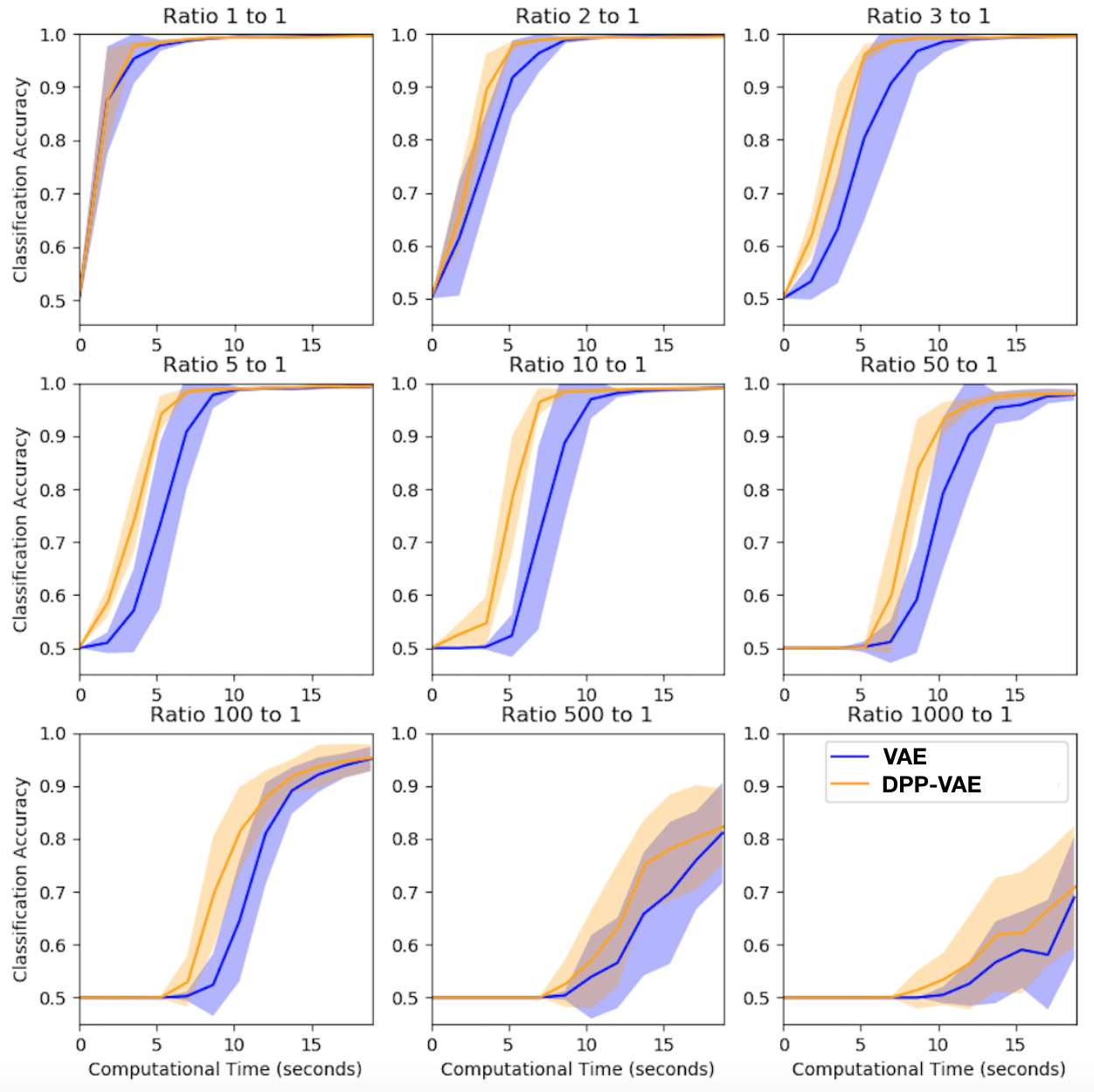} 
\caption{Comparing classification accuracy on balanced test data against computational time, as the imbalance ratio of training data varies.}
\label{fig:acc_compare_time_ratio}
\end{figure}

\paragraph{Classification}
We first compare standard VAE and DPP-VAE on a binary classification task based on MNIST data. The training examples include 5000 MNIST `0', `1' handwritten digits data. For the balanced case, there are 2500 Class 0 and 2500 Class 1 observations. We then vary the imbalance ratio from 1:1 to 1:1000, where digit `1' is considered as the minor class. The test set is a balanced dataset with 500 Class 0 and 500 Class 1 observations.

We select the latent dimension to be 20 and set $(\alpha, \rho, \sigma) = (1000, 1, 1)$. Both methods are trained for 10 epochs with the batch size set to 100. Monte Carlo sample of size 1 is used since the batch size is large enough. A two layer convolutional and deconvolutional neural network with the ReLU activation function is used for encoding and decoding the images. The latent samples are then used for classification. To this end, we use the learned features in a simple logistic regression model with optimal hyperparameters selected through cross-validation.

The performance of the two methods on balanced test sets is shown in Figure \ref{fig:acc_compare_time_ratio}. The results are averaged over 20 independent runs. As we can see, for a given computation time, the DPP-VAE model tends to achieves a significantly higher classification accuracy rate compared to the standard VAE. The improvement becomes more obvious as the imbalance ratio increases.

\begin{table}[!b]
	\caption{Generated minor class (digit `1') percentage}
	\vspace{10pt}
	\label{table:class_ratio}
	\centering
	\begin{tabular}{llll}
	    \toprule
		Class ratio & Training (\%)& VAE (\%) & DPP-VAE (\%)\\
		\midrule
		1:10 & 9.1\% &  7.2\% & \textbf{17.7\%}  \\
		1:100 & 1\% & 1.21\% &  \textbf{3.68\%}  \\
		1:1000 & 0.0999\% & 0.0562\% &  \textbf{0.9469\%} \\
		\bottomrule
	\end{tabular}
\end{table}

\paragraph{Balancing data generation}
To further investigate our method, we use both VAE and DPP-VAE to generate digits under three different scenarios: 1:10 ratio, 1:100 ratio, and 1:1000 ratio. Here, the parameter settings are the same as before.

Synthetic MNIST data (digit `0' and `1') are generated by standard VAE and DPP-VAE separately. For this, random latent vectors from standard normal distribution are fed into the trained decoder network to generate handwritten `0' and `1' digits. The same latent vectors are used for both methods. The total training data is set to 5000.

Results are displayed in Table \ref{table:class_ratio}. As we can see, the percentage of the minor class is substantially increased for our model, and the effect is more significant as the imbalance ratio increases: 1.9 times increase for 1:10 ratio (17.7\% versus 9.1\%), 3.7 times increase for 1:100 ratio (3.68\% versus 1\%), and 9.5 times increase for 1:1000 ratio (0.9469\% versus 0.0999\%). This shows that our method is up-weighting the minor class.

\subsection{Neural decoding}
We now apply our proposed method to the odor experiment, which was discussed above and was the original inspiration for this research. While most existing methods for class imbalance focus on binary classification problems, here we can show that our method can be easily applied to multi-class problems, which are more challenging in general. We also show that our proposed method can improve latent representation learning by avoiding redundancy in the latent space. 

As discussed above, in our experiment, rats perform a sequence memory task and are expected to recognize five different odors presented in a specific order: A, B, C, D, and E. Here, we are interested in decoding their corresponding neural spike signals for odor classification. Neural decoding refers to the mapping from neural activities to stimulus (here, odors). In our experiment, a sequence of trials always starts with presenting odor A and terminates early once the rat makes a mistake. Therefore, there are more class A trails than other classes, and decoding results are highly affected by this imbalance. More specifically, there are 58 trials for odor A, 41 trials for odor B, 37 trials for odor C, 32 trials for odor D, and 26 trials for odor E. Previous studies have shown that a given odor is most related to the neural activities occurring during the 0.15s-0.4s time window after odor presentation. Thus, neural spike data during that time window are used for training both VAE and DPP-VAE models along with the multinomial logit model that uses the latent features as predictors.

\begin{table}[!b]
	\caption{Comparing VAE and DPP-VAE in terms of Precision (P), Recall (R), and F1 Score using 6-fold cross-validation.}
	\label{table:neuron_cv}
		\centering
		\begin{tabular}{llll | lll}
		& \multicolumn{3}{c}{VAE} & \multicolumn{3}{c}{DPP-VAE} \\
		\toprule
			&  P & R &  F1 &  P & R &  F1\\
			\midrule
			A &  0.706 & 0.875 & 0.776 &  0.751 & 0.917 & \textbf{0.809} \\
			B &  0.636 & 0.625 & \textbf{0.621} &  0.497 & 0.708 & 0.575 \\
			C &  0.233 & 0.417 & 0.294 &  0.361 & 0.458 & \textbf{0.377 } \\
			D &  0.215 & 0.167 & 0.172 &  0.333 & 0.250 & \textbf{0.278} \\
			E &  0.139 & 0.083 & 0.103 & 0.333 & 0.083  & \textbf{0.133}  \\
			avg &  0.386 & 0.433 & 0.393 &  0.455 & 0.483 & \textbf{0.434}  \\
			\bottomrule
		\end{tabular}
\end{table}

%
%
\begin{figure}[!ht]
\centering
\includegraphics[width=0.56\textwidth]{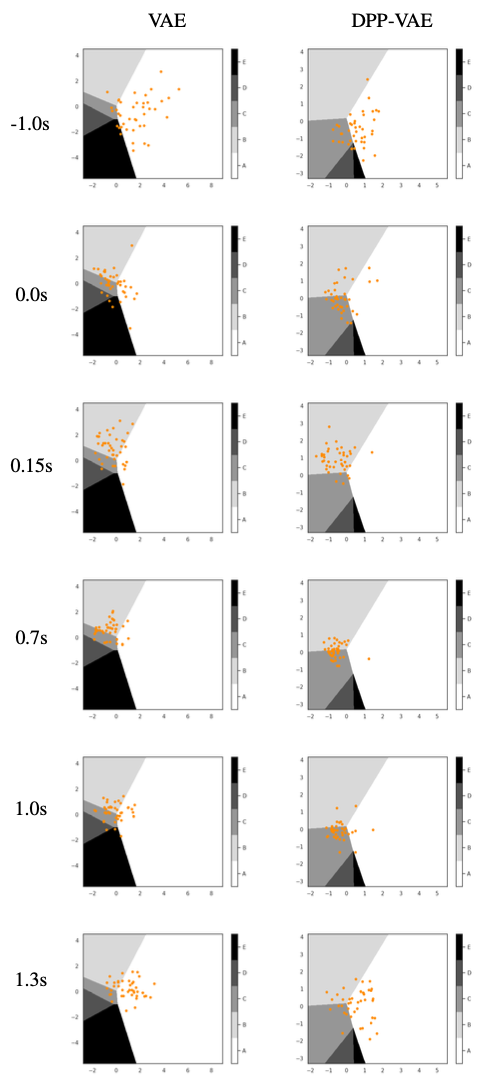}
	\caption{Latent representation of the decoding process using VAE and DPP-VAE during the presentation of odor B. Each row corresponds to a specific time point starting from 1 second before the odor presentation (denoted as time 0). Note that each complete trial lasts about 1.2s. Each point represents one trial. As we can see, our model clearly shows the movement of trials towards region C (i.e., the upcoming odor) around 0.7s after odor B was presented.}\label{fig:neuralLatentSpaceB}
\end{figure}

\paragraph{Classification results} The average performance based on a 6-fold cross-validation is displayed in Table \ref{table:neuron_cv}.
The test data for each fold is selected to be a balanced set, with 4 trials for each class. As the results show, our proposed method improves the performance (e.g., F1 score) of the minor classes (C, D and E), without negatively affecting the performance on the major class, odor A (although, the accuracy rate for odor B dropped slightly). The overall performance has also improved substantially.

\paragraph{Visualization for sequence memory replay} Besides neural decoding, another fundamental question that is of immense interest in neuroscience is \emph{sequence memory replay}. That is, in our experiment we are interested to find out whether the neural activities corresponding to the five odors are replayed in the sequence of ABCDE during each session. Here, we use our method to investigate this phenomenon for the designed nonspatial task, which is quite different from typical experiments discussed in the literature. 

The general framework, as before, is to use neural activities during the 0.15-0.4s time window relative to the nose-poke as our training data, and predict/reconstruct odors using neural activities during other time windows within each trial. We expect to identify some sort of replay during the -2s to 2s time frame, regarding the odor presentation as time 0. The neural activities outside of the 0.15-0.4s window are treated as the test data. 

More specifically, we first train VAE and DPP-VAE using the 0.15-0.4s neural data as before. We keep the latent dimension to be 20 since VAE works better with a moderate dimension. Then, for the purpose of visualization, we use the first two principal components of the latent representation to train a multinomial logit model. Note that applying these linear models to latent variables is reasonable since we expect that the decoding process in VAE and DPP-VAE maps nonlinear patterns in the original space to relatively linear patterns in the latent space. The decision boundaries of the classification model can then be visualized using a 2-dimensional plot. 

We focus on replay pattern during trials associated with odor B presentation. To investigate sequence replay, we simply examine different time windows such as $[-2s, -1.75s]$, $\cdots$, $[0s, 0.25s]$, $\cdots$, $[1.9s, 2.15s]$ for all B trials. Within each time window, the 2-dimensional representation for all odor B trials are visualized in the same plot. The representations in the test windows are generated by using the models that are trained on the 0.15-0.4s window.

Figure \ref{fig:neuralLatentSpaceB} shows the results based on VAE and DPP-VAE for a subset of windows during the presentation of odor B. The latent space is divided into different regions based on the decision boundaries obtained from the multinomial logit model. Each point on the plot corresponds to one trial projected into the latent space. As we can see, before the odor presentation (-1s), the trails are randomly scattered. The movement towards region B starts around 0s (at the time of odor presentation) indicating the rat's anticipation for the upcoming odor. Most trails move into region B around 0.15s, which is expected since this is the training window. However, the most interesting part of our results is what happens right after that around 0.7s, which is a test window: the trials clearly move into region C, which is the next odor in the sequence while the rat is still in trial B (recall that the trial ends after 1.2s). This is an evidence of sequence replay, which has not been shown for this type of nonspatial experiments in the past. More importantly, note that while the overall neural patterns throughout the course of a trial are somehow similar between VAE and DPP-VAE, the patterns are more clear using our proposed DPP-VAE model: around 1s after the odor presentation, our method shows that most trials have moved into C region.

\section{Discussion}
In this work, we have proposed a novel latent representation learning method based on diversity-encouraging Determinantal Point Processes to alleviate the class imbalance problem and to provide a more clear latent representation of the data. To this end, we have modified the standard variational auto-encoder method by using continuous k-DPP as the prior for latent variables. Further, we have developed the required inference algorithm to implement this model. Using synthetic and real data, we have shown that our proposed method can in fact improve latent representation learning, as well as the prediction accuracy of classification models (especially for minor classes). Our work can have significant contributions to many areas of machine learning, especially when samples of rare classes with great importance are hard to collect. Additionally, our proposed method can lead to finding novel phenomena in the field of neuroscience by providing unprecedented insight into underlying structure of neural data. 


\section*{Acknowledgement}
This work is supported by NSF grant DMS 1622490 and NIH grant R01 MH115697.

\bibliographystyle{unsrt}

\end{document}